\documentclass[twoside,twocolumn,9pt]{extarticle}

\usepackage{blindtext} 

\usepackage{mathptmx}
\usepackage[T1]{fontenc} 
\usepackage{microtype} 

\usepackage{graphicx} 
\usepackage{flafter}
\usepackage[english]{babel} 

\usepackage[a4paper,left=0.71in,top=0.98in,right=0.71in,bottom=0.98in,columnsep=15pt]{geometry} 
\usepackage[hang, small,labelfont=bf,up,textfont=it,up]{caption} 
\usepackage{booktabs} 

\usepackage{enumitem} 
\setlist[itemize]{noitemsep} 

\usepackage[runin]{abstract} 
\abslabeldelim{:}

\usepackage{titlesec} 
\titleformat{\section}[block]{\large\bfseries}{\thesection.}{1em}{} 
\titleformat{\subsection}[block]{\large}{\thesubsection.}{1em}{} 
\titleformat{\subsubsection}[block]{\normalsize}{\thesubsubsection.}{1em}{} 

\usepackage{titling} 

\usepackage{authblk}

\usepackage{caption}
\DeclareCaptionLabelFormat{nospace}{#1#2}
\usepackage[utf8]{inputenc}   				 	
\usepackage{silence}  							

\usepackage{url}
\usepackage{letltxmacro}
\LetLtxMacro{\autocite}{\cite}
\LetLtxMacro{\textcite}{\cite}

\usepackage[fleqn]{amsmath}
\usepackage{amsthm}
\usepackage{amsfonts}
\usepackage[capitalize]{cleveref}
\crefname{equation}{}{}
\usepackage{xcolor}
\usepackage{xparse,xstring}
\usepackage{subcaption}

\usepackage{tikz}
\usetikzlibrary{shapes}
\usepackage{pgfplots}
\newlength\figurewidth
\newlength\figureheight
\newlength\venncircle
\newlength\realworldwidth
\newlength\realworldheight

\theoremstyle{plain}
\newtheorem{definition}{Definition}

\newcommand{\scenario}{s}
\newcommand{\scenarioa}{a}
\newcommand{\scenariob}{b}
\newcommand{\scenarioc}{c}
\newcommand{\scenariocategory}{S}
\newcommand{\scenariocategorya}{A}
\newcommand{\scenariocategoryb}{B}
\newcommand{\scenariocategoryc}{C}

\providecommand{\keywords}[1]{\textbf{KEY WORDS:} #1}

\pretitle{\begin{center}\huge\normalfont\MakeUppercase} 
\posttitle{\end{center}\vskip 1em} 
\title{Tagging Real-World Scenarios for the Assessment of Autonomous Vehicles} 

\author[1,2]{\large\bfseries Erwin de Gelder}
\author[1]{\large\bfseries Olaf Op den Camp}

\affil[1]{\normalsize\textit{TNO, Integrated Vehicle Safety, Helmond, The Netherlands (E-mail: erwin.degelder@tno.nl, olaf.opdencamp@tno.nl)}}
\affil[2]{\normalsize\textit{Delft University of Technology, Delft Center for Systems and Control, Delft, The Netherlands}}
\date{} 
\renewcommand{\maketitlehookd}{%
\begin{abstract}	
\noindent The development of Autonomous Vehicles (AVs) has made significant progress in the last years and it is expected that AVs will soon be introduced on our roads. An essential aspect in the development of AVs is the assessment of quality and performance aspects of the AVs, such as safety, comfort, and efficiency. Among other methods, a scenario-based approach has been proposed. With scenario-based testing, the AV is subjected to a collection of scenarios that represent real-world situations. 

The collection of scenarios needs to cover the variety of what an AV can encounter in real traffic. As a result, many different scenarios are considered, that are grouped into so-called scenario categories. We propose a method for defining the scenario categories using a system of tags, where each tag describes a particular characteristic of a scenario category.

There is a balance between having generic scenario categories – and thus a high variety among the scenarios in the scenario category – and having specific scenario categories without much variety among the scenarios in the scenario category. For some systems, one is interested in very specific set of scenarios, while for another system one might be interested in a set of scenarios with a high variety. To accommodate this, tags are structured in trees. The different layers of the trees can be regarded as different abstraction levels.

Next to presenting the method for describing scenario categories using tags, we will illustrate the method by showing applicable trees of tags using concrete examples in the Singapore traffic system. Trees of tags are shown for the vehicle under test, the dynamic environment (e.g., the other road users), the static environment (e.g., the road layout), and the environmental conditions (weather and lighting conditions). Few examples are presented to illustrate the proposed method for defining the scenario categories using tags.

\keywords{autonomous vehicle, assessment, scenarios}
\end{abstract}
}

\begin{document}

\maketitle


\section{Introduction}
\label{sec:introduction}

An essential aspect in the development of autonomous vehicles (AVs) is the assessment of AVs \autocite{bengler2014threedecades, stellet2015taxonomy, Helmer2017safety, putz2017pegasus, roesener2017comprehensive, gietelink2006development, wachenfeld2016release}.
For legal and public acceptance of AVs, a clear definition of system performance is important, as well as quantitative measures for system quality. 
Traditional methods, such as \autocite{response2006code, ISO26262}, used for the evaluation of driver assistance systems, are no longer sufficient for the assessment of quality and performance aspects of an AV, because they would require too many resources \autocite{wachenfeld2016release}. 
Therefore, among other methods, a scenario-based approach has been proposed \autocite{elrofai2018scenario, putz2017pegasus}. 

\begin{figure*}[t]
	\centering
	\includegraphics[width=\linewidth]{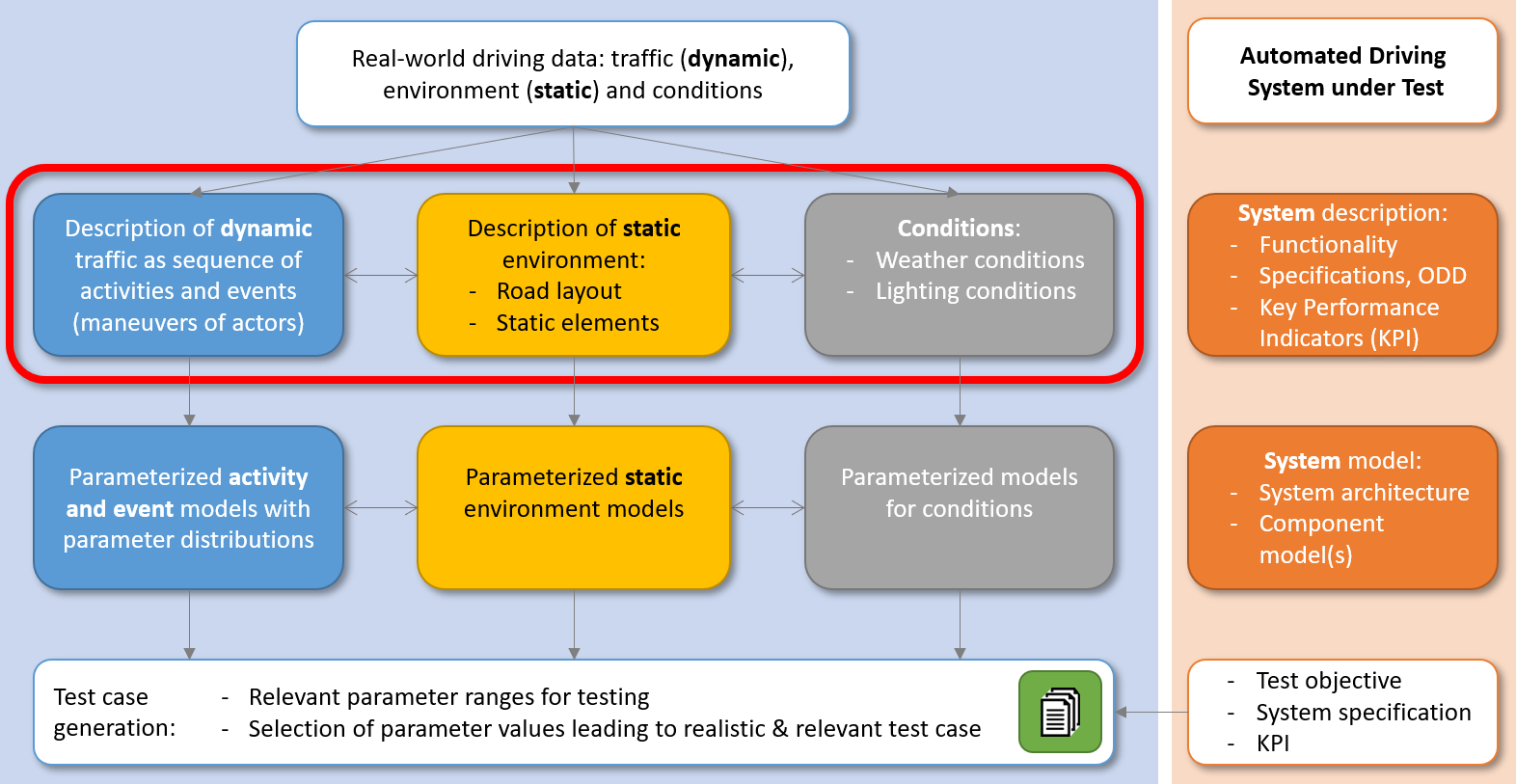}	
	\caption{A schematic overview of the required components to describe a scenario, and the relation to AV specifications and test cases, based on \cite{elrofai2018scenario}. The red box indicates the focus of the current report.}
	\label{fig:scenario schematic}
\end{figure*}

It is important that the collection of scenarios used for the scenario-based assessment of AVs cover the variety of what an AV can encounter during real operation in traffic. 
\Cref{fig:scenario schematic} provides a schematic overview of the required components to describe the test cases that an AV should be subjected to \autocite{elrofai2018scenario}. 
Two different abstraction levels can be considered where the first level qualitatively describes the scenarios and the second level quantitatively describes the scenarios using parameters and models \autocite{degelder2020ontology}. In a similar manner, the AV can be described, such that the relevant test cases can be deduced. 

Because the scenarios need to cover the variety of what an AV can encounter in real traffic, many different scenarios need to be considered. To handle a large number of scenarios, this paper focuses on the qualitative description of the scenarios. Therefore, the scenarios are categorized into so-called scenario categories, where a scenario category can be regarded as an abstraction of a quantitative scenario \autocite{degelder2020ontology}. For example, ``Lead vehicle braking'' is a scenario category referring to the scenarios in which a car in front of the ego vehicle brakes. 

In this paper, we propose a method for defining the scenario categories using a system of tags. Next to proposing the method, we introduce an appropriate system of tags to describe different scenario categories that cover a large portion of the possible varieties that are found in real-world traffic.
It is the objective to get a fair indication of the safe operation of the AV when deployed in real-world traffic by subjecting the AV to test cases (both in physical tests, e.g., on a test track, and virtual simulations using appropriate AV models) based on the scenario categories that are defined using the system of tags proposed in this paper.

This paper does not provide an overview of required test cases. The test cases, however, might be based on the scenario categories that are defined using the presented tags. Before selecting and generating test cases, a match needs to be made of the AV's Operational Design Domain (ODD) \autocite{sae2018j3016} onto the requirements for deployment of the AV in a certain area. 
Based on the ODD, test cases can be selected for the safety assessment of the AV according to, e.g., ISO~34502 \autocite{ISO34502}. Describing a process for selecting test cases is out of the scope of the current paper. 

This paper is organized as follows. In \cref{sec:scenario category}, we explain what we mean with a scenario category and how this relates to scenarios and test cases. Next, in \cref{sec:tags}, we propose a selection of tags, structured in trees, to define scenario categories. Few examples of scenario categories defined using the tags that are proposed in \cref{sec:tags} are presented in \cref{sec:examples}. We conclude this paper in \cref{sec:conclusions}.

\section{Terminology}
\label{sec:scenario category}

In this section, we first explain what we mean with scenarios and scenario categories. Next, in \cref{sec:tags for scenario catgories}, we describe why tags are used to define scenario categories. We end this section in \cref{sec:test cases} with an explanation of the relation between the scenario categories and test cases for the assessment of AVs.

\subsection{What is a scenario category?}
\label{sec:definition scenario category}

We distinguish quantitative scenarios from qualitative scenarios, using the definitions of \emph{scenario} and \emph{scenario category} of \autocite{degelder2020ontology}:

\begin{definition}[Scenario]
	\label{def:scenario}
	A scenario is a quantitative description of the relevant characteristics of the ego vehicle, its activities and/or goals, its static environment, and its dynamic environment. In addition, a scenario contains all events that are relevant to the ego vehicle.
\end{definition}

\begin{definition}[Scenario category]
	\label{def:scenario category}
	A scenario category is a qualitative description of the ego vehicle, its activities and/or goals, its static environment, and its dynamic environment.
\end{definition}

Introducing the concept of scenario categories brings the following benefits \autocite{degelder2020ontology}:
\begin{itemize}
	\item For a human, it is easier to interpret a qualitative description rather than a quantitative description.
	\item It enables to refer to a group of scenarios that have something in common. Therefore, it enables characterization of the type of scenarios, thus making discussing scenarios much easier.
	\item It allows for quantifying the completeness of a set of scenarios by separately quantifying the completeness of scenario categories and the completeness of scenarios in each category.
	This is easier because scenario categories are discrete by nature whereas scenarios are continuous. See \autocite{degelder2019completeness} for more details.
\end{itemize}

To describe the relation between a scenario and a scenario category, the verb ``to comprise'' is used \autocite{degelder2020ontology}.
If a specific scenario category $\scenariocategory$ is an abstraction of a specific scenario $\scenario$, then we say that the specific scenario category $\scenariocategory$ comprises that specific $\scenario$. 
To describe the relation between two scenario categories, the verb ``to include'' is used \autocite{degelder2020ontology}. 
To further illustrate this, consider \cref{fig:venn diagram scenario category}, where $\scenarioa$, $\scenariob$, and $\scenarioc$ represent scenarios and $\scenariocategorya$, $\scenariocategoryb$, and $\scenariocategoryc$ represent scenario categories.
The following principles apply for these relations:
\begin{itemize}
	\item A given scenario category can comprise multiple scenarios, e.g., in \cref{fig:venn diagram scenario category}, $\scenariocategoryb$ comprises $\scenarioa$ and $\scenariob$.
	\item Multiple scenario categories can comprise a specific scenario, e.g., in \cref{fig:venn diagram scenario category}, both $\scenariocategorya$ and $\scenariocategoryb$ comprise $\scenariob$. 
	\item A scenario category includes another scenario category if it comprises all scenarios that the other scenario category comprises, e.g., in \cref{fig:venn diagram scenario category}, $\scenariocategoryb$ includes $\scenariocategorya$.
\end{itemize}

\setlength{\venncircle}{6em}
\begin{figure}[h]
	\centering
	
	\begin{tikzpicture}
		\fill[red!50, fill opacity=0.5] (-\venncircle/2, 0) circle (\venncircle);
		\fill[green!50, fill opacity=0.5] (\venncircle/2, 0) circle (\venncircle);
		\draw (-\venncircle/2, 0) circle (\venncircle);
		\draw (\venncircle/2, 0) circle (\venncircle);
		
		\node[anchor=east](daylight) at (-5/4*\venncircle, \venncircle) {$\scenariocategoryb$};
		\draw (daylight) -- ({(-sqrt(2)/2-1/2)*\venncircle}, {sqrt(2)/2*\venncircle});
		\node[anchor=west](rain) at (5/4*\venncircle, \venncircle) {$\scenariocategoryc$};
		\draw (rain) -- ({(sqrt(2)/2+1/2)*\venncircle}, {sqrt(2)/2*\venncircle});
		\node[anchor=south](day and rain) at (-1/4*\venncircle, 8/7*\venncircle) {$\scenariocategorya$};
		\draw (day and rain) -- ({(1/2-sqrt(2)/2)*\venncircle}, {(sqrt(2)/2)*\venncircle});
		
		\node at (-0.1*\venncircle, -0.1*\venncircle) (scenariob){\textbullet};
		\node at (-\venncircle, 0.3*\venncircle) (scenarioa){\textbullet};
		\node at (\venncircle, 0.2*\venncircle) (scenarioc){\textbullet};
		\node[right of=scenarioa, node distance=1em]{$\scenarioa$};
		\node[right of=scenariob, node distance=1em]{$\scenariob$};
		\node[right of=scenarioc, node distance=1em]{$\scenarioc$};
	\end{tikzpicture}

	\caption{The red and green circles correspond to the scenario categories $\scenariocategoryb$ and $\scenariocategoryc$, respectively. The overlap between the two circles corresponds to scenario category $\scenariocategorya$. The dots represent scenarios $\scenarioa$, $\scenariob$, and $\scenarioc$.}
	\label{fig:venn diagram scenario category}
\end{figure}
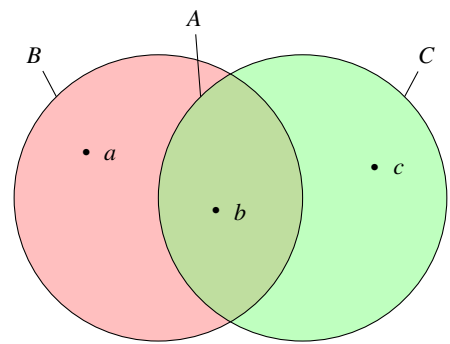

\subsection{Why using tags?}
\label{sec:tags for scenario catgories}

Tags are often used when providing extra information on a piece of data \autocite{smith2007tagging}. A tag is a keyword or a term that helps describing an item. For example, items in a database can contain some tags that enable users to quickly obtain several items that share a certain characteristic described by a tag \autocite{craft2004tagging}. 
Applications are very broad, e.g., from classification of audio data \autocite{kong2017joint} capturing musical characteristics from songs \autocite{ellis2011semantic} to tagging of Wikipedia pages \autocite{voss2006collaborative}.
The use of these tags brings several benefits:
\begin{itemize}
	\item The tags of a scenario can be helpful in determining which scenario categories do and do not comprise the scenario \autocite{degelder2020scenariomining}.
	\item It is easy to select scenarios from a scenario database or a scenario library by using tags or a combination of tags.
	\item As opposed to the proposed categorization of scenarios in \autocite{opdencamp2014cats, USDoT2007precrashscenarios, lenard2014typical, lara2019harmonized, lara2019harmonized}, scenario categories do not need to be mutually exclusive.
\end{itemize}

There is a balance between having generic scenario categories --- and thus a wide variety among the scenarios belonging to the scenario category --- and having specific scenario categories without much variety among the scenarios in the scenario category. For some systems, one may be interested in very specific set of scenarios, while for another system one might be interested in a set of scenarios with a high variety. To accommodate this, tags can be structured in hierarchical trees \autocite{molloy2017dynamic}. The different layers of the trees can be regarded as different abstraction levels \autocite{Bonnin2014}.

\subsection{Test cases}
\label{sec:test cases}

Scenario categories can be used to define relevant test cases for the assessment of AVs, see, e.g., \cite{ploeg2018cetran,elrofai2016scenario}. It is generally acknowledged that test cases for the safety assessment of AVs should be based on real-world scenarios \cite{putz2017pegasus, roesener2016scenariobased, deGelder2017assessment}. Nevertheless, the terms scenario and test case are often confused; also the combination test scenario is often used in discussions\footnote{We consider a \emph{test scenario} to be similar to \emph{test case}, but we prefer using the term \emph{test case} as to reduce confusion with the term \emph{scenario}.}. 

We use the term scenario for a description of a situation that can happen or has happened in the real world. In other words, scenarios are used to describe any type of situation that a vehicle in operation can encounter during its lifetime. The set of scenarios described by the scenario categories will not fully cover all possible situations that can occur in reality. In \cref{fig:test cases}, this is represented by the fact that the available set of scenarios (red) does not cover all relevant situations in the real world (blue).

\definecolor{realworld}{RGB}{50, 50, 192}
\definecolor{scenarios}{RGB}{192, 50, 50}
\definecolor{ODD}{RGB}{50, 192, 50}
\definecolor{testcases}{RGB}{255, 255, 153}
\setlength{\realworldwidth}{15em}
\setlength{\realworldheight}{9em}
\tikzstyle{thickness}=[ultra thick]
\begin{figure}[t]
	\centering
	\begin{tikzpicture}	
		\begin{scope}
		\clip (.1\realworldwidth, 0) ellipse (.7\realworldwidth/2 and .8\realworldheight/2); 
		\fill[testcases] (-.1\realworldwidth, 0) ellipse (.7\realworldwidth/2 and .8\realworldheight/2);
		\end{scope}
		
		\node[ellipse, minimum height=\realworldheight, minimum width=\realworldwidth, realworld, draw, thickness](real world) at (0, 0) {};
		\node(real world text) at (.45\realworldwidth, .5\realworldheight) {Real world};
		\draw[thickness, realworld](real world) -- (real world text);
		
		\node[ellipse, minimum height=0.8\realworldheight, minimum width=.7\realworldwidth, scenarios, draw, thickness](scenarios) at (-.1\realworldwidth, 0) {};
		\node(scenarios text) at (-.45\realworldwidth, .5\realworldheight) {Scenarios};
		\draw[thickness, scenarios](scenarios) -- (scenarios text);
		
		\node[ellipse, minimum height=.8\realworldheight, minimum width=.7\realworldwidth, ODD, draw, thickness](odd) at (.1\realworldwidth, 0) {};
		\node(odd text) at (.45\realworldwidth, -.45\realworldheight) {ODD};
		\draw[thickness, ODD](odd) -- (odd text);
		\node at (0, 0) {Test cases};
	\end{tikzpicture}
	\caption{The relation between the real world, the ODD, the scenarios, and test cases.
	}
	\label{fig:test cases}
\end{figure}
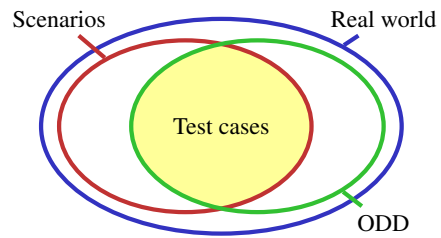

To describe the difference between a scenario and a test case, it is important to know a vehicle's Operational Design Domain (ODD)\footnote{SAE automotive standard J3016 \autocite{sae2018j3016} defines an ODD as ``operating conditions under which a given driving automation system or feature thereof is specifically designed to function, including, but not limited to, environmental, geographical, and time-of-day restrictions, and/or the requisite presence or absence of certain traffic or roadway characteristics.''}, to determine the set of relevant test cases. The ODD depends on the application of an AV and usually is the result of the design of the AV in relation to the requirements of the application. An ODD covers a dedicated and limited area of the real world as indicated in \cref{fig:test cases} by the green set.


Once the ODD is known, and we have scenarios that cover (part of) the ODD, test cases can be generated. The set of test cases is considered to be a subset of scenarios, as not all scenarios are relevant for each type of vehicle or each type of application. However, test cases are always generated from scenarios; we therefore assume that no test case is generated in areas not covered by the set of scenarios. It is unnecessary to provide test cases outside the ODD, as the system is not expected to respond outside the ODD. The set of test cases in \cref{fig:test cases}, is denoted by the yellow shaded area. It is represented by the intersection of the set of scenarios and the ODD. In the ideal case, the scenario set is complete and encompasses the ODD. In \cref{fig:test cases}, this would show if the ODD (green) would fully fit within the set of scenarios (red).

\section{Selection of tags}
\label{sec:tags}

The definition of tags and trees of tags will be presented subsequently for the dynamic environment, for the static environment, and finally for the conditions.

\begin{figure*}[t!]
	\centering
	\includegraphics{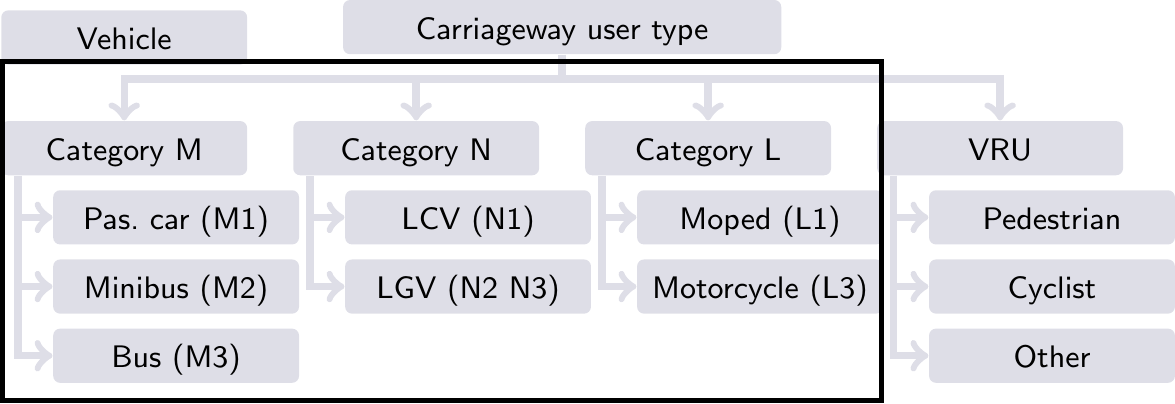}
	\caption{Tags for the type of the carriageway user, with a reference to UNECE vehicle categories \autocite{UNECE2011consolidated}. 
		A vehicle of category M, N, or L is considered as a \emph{vehicle} in the context of this report as indicated by the black box outline. 
	}
	\label{fig:tree carriageway user type}
\end{figure*}

\begin{figure*}[t!]
	\centering
	\includegraphics{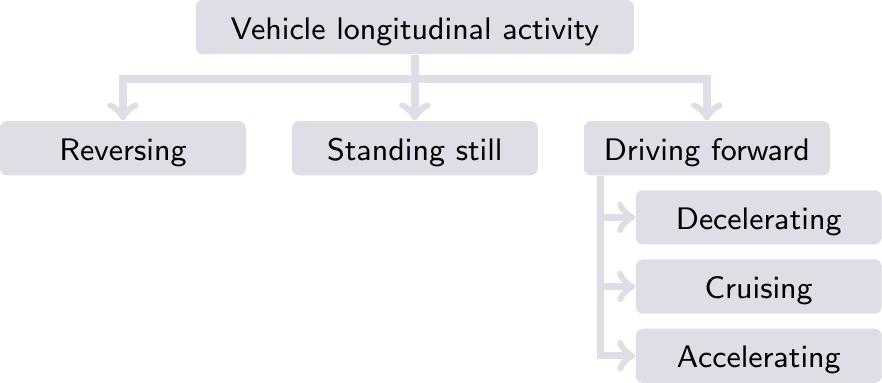}\\
	\vspace{0.5em}
	\includegraphics{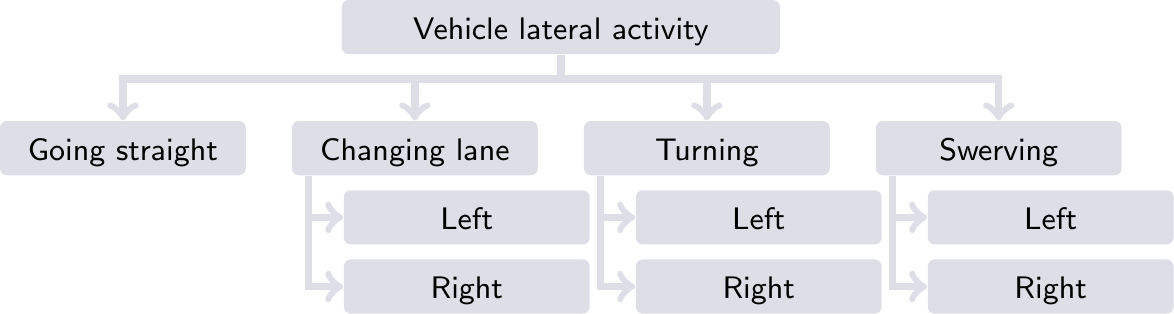}
	\caption{Tags for lateral and longitudinal activities of a vehicle. 
	}
	\label{fig:activities}
\end{figure*}

\subsection{Tags for the dynamic environment}
\label{sec:selection of tags dynamic}

To describe the dynamic environment, the activities of the different actors are described. First, we consider different type of actors in \cref{sec:type of actor}. Next, in \cref{sec:activities}, tags are provided to describe the activities of the actors. In \cref{sec:initial state}, tags are presented that describe the initial state of an actor in a scenario. Some special tags are applicable for vehicle driving in front of the ego vehicle, see \cref{sec:lead vehicle}. 

\subsubsection{Carriageway user type}
\label{sec:type of actor}

A first distinction within a scenario is usually made for the type of carriageway user, see \cref{fig:tree carriageway user type}. The tree of tags is not considered to be complete, however, the current tags cover the most common type of carriageway users. For the motorized vehicle a reference is made to the UNECE regulation \autocite{UNECE2011consolidated}. In the regulation, a further distinction in vehicle categories is made. The more general tag ``vehicle'' applies if a vehicle could be either of category M, N, or L. For the category of Vulnerable Road Users (VRU), the European convention is used, with the exception that powered two wheelers, such as a motorcycle, are explicitly considered a vehicle and not a VRU. The reason to use the separate category L, i.e., motor vehicles with less than four wheels, is the large difference in behavior they exhibit compared to VRU; their position on the road and their riding dynamics including speed are just two of the striking differences. 


\subsubsection{Activities}
\label{sec:activities}

An activity describes the behavior related to an actor. This includes, but is not limited to, the dynamic driving tasks as mentioned in SAE J3016 \cite{sae2018j3016}. In this paper, only the lateral motion control (via steering) and longitudinal motion control (via acceleration and deceleration) are reflected into tags.

The lateral and longitudinal activities of the a vehicle are characterized by the tags of \cref{fig:activities}. The tags may also refer to the objective of the ego vehicle in case no activities are defined. For example, a test case in which the ego vehicle's objective is to make a left turn, the tags ``Turning'' and ``Left'' are applicable. 


Four different types of activities are identified regarding the lateral movement. Here, it is assumed that ``Lateral'' refers to the direction perpendicular to the lane the vehicle is driving in. Therefore, if the vehicle is driving on a curved road while staying more or less in its lane (lane following), the tag ``Going straight'' is applicable. When the vehicle changes lane to an adjacent lane, the tag ``Changing lane'' is applicable. The tag ``Turning'' is applicable when the vehicle turns at a junction. The tag ``Swerving'' is applicable when the vehicle significantly changes lateral position without performing a complete lane change. For example, when the vehicle overtakes a cyclist that is riding at one side of the lane, the vehicle might swerve to the other side of the lane. 

Three different types of activities are identified regarding the longitudinal movement. A distinction is made between driving forward, reversing, and standing still. Regarding driving forward, a further distinction is made with respect to the acceleration.

Due to the typical dynamics for pedestrians and cyclists, separate tag trees are envisioned to characterize their behavior. However, for the sake of brevity, the actual trees are omitted here. We refer the interested reader to \autocite{degelder2019scenariocategories} where we also present these tag trees.

\begin{figure*}[t!]
	\centering
	\includegraphics{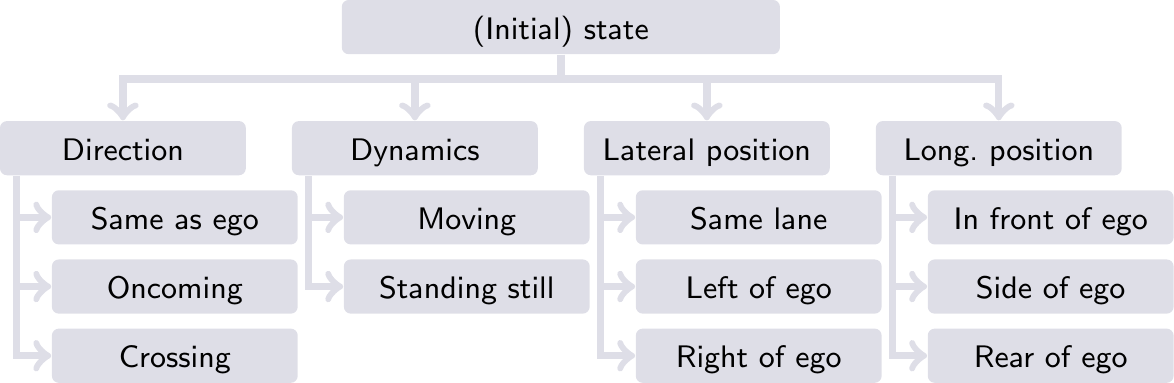}
	\caption{Tags regarding the initial position and movement of road users in a scenario.}
	\label{fig:tree initial state}
\end{figure*}

\subsubsection{State or initial state}
\label{sec:initial state}

\Cref{fig:tree initial state} shows tags for the state or the initial state of the potential other road users with respect to the ego vehicle. A distinction is made in the direction of orientation of the road user, the dynamics, and the longitudinal and lateral position. Three tags, i.e., ``Same as ego'', ``Oncoming'', and ``Crossing'', refer to the direction of the road user with respect to the direction of the ego vehicle. 
The tag ``Dynamics'' distinguishes between moving and standing still. 
Finally, two tags are used to describe the position of the actor with respect to the ego vehicle, in longitudinal and lateral direction.

\subsubsection{Lead vehicle}
\label{sec:lead vehicle}

\begin{figure}
	\centering
	\includegraphics{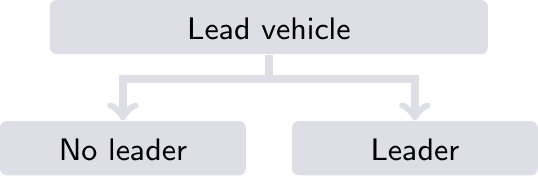}
	\caption{Tags for a lead vehicle, i.e., a vehicle driving in front of ego vehicle.}
	\label{fig:tree lead vehicle}
\end{figure}

\Cref{fig:tree lead vehicle} contains the tags that are used to mark specific actors as either being a leading vehicle or not.

\begin{figure}
	\centering
	\includegraphics{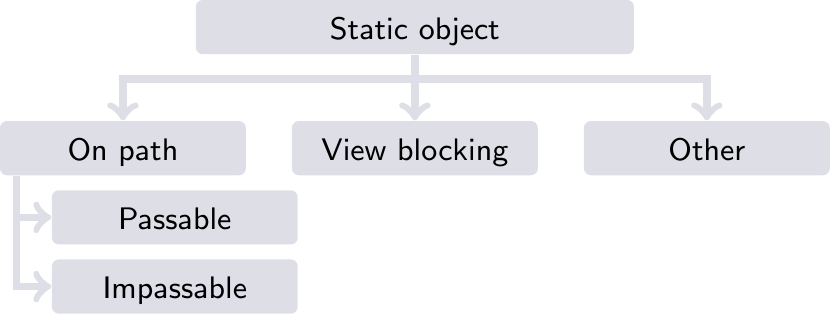}
	\caption{Tags that describe a static object.}
	\label{fig:tree static object}
\end{figure}

\begin{figure*}[t!]
	\centering
	\includegraphics{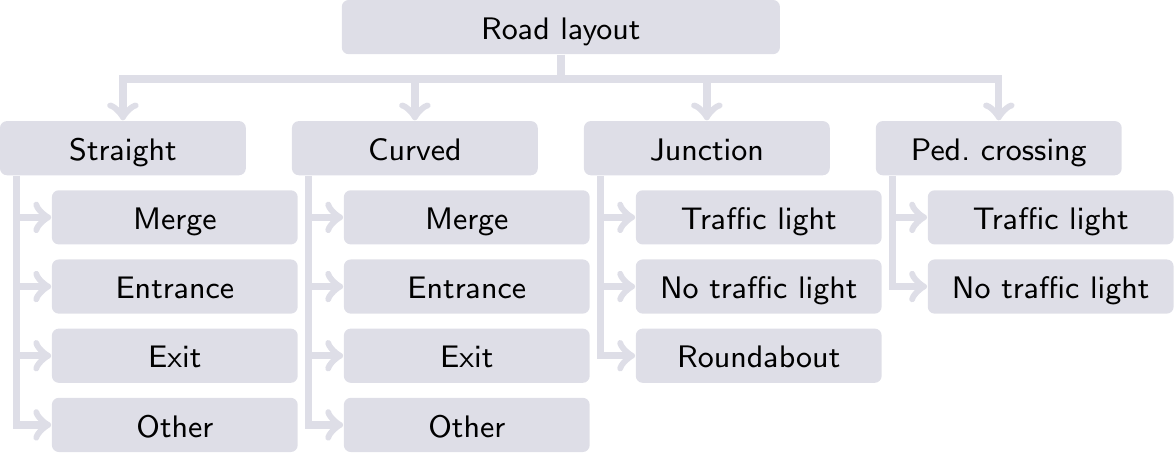}
	\caption{Tags that describe the road layout.}
	\label{fig:tree road layout}
\end{figure*}

\begin{figure*}[t!]
	\centering
	\includegraphics{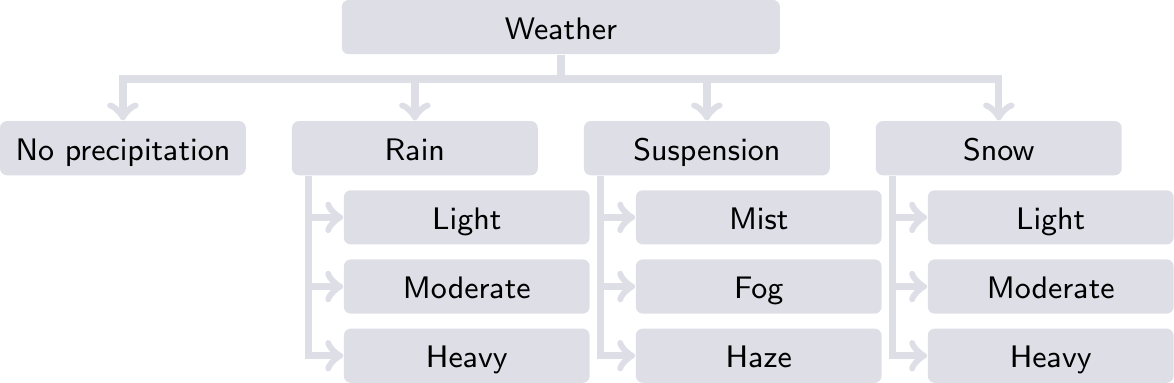}\\
	\vspace{0.5em}
	\includegraphics{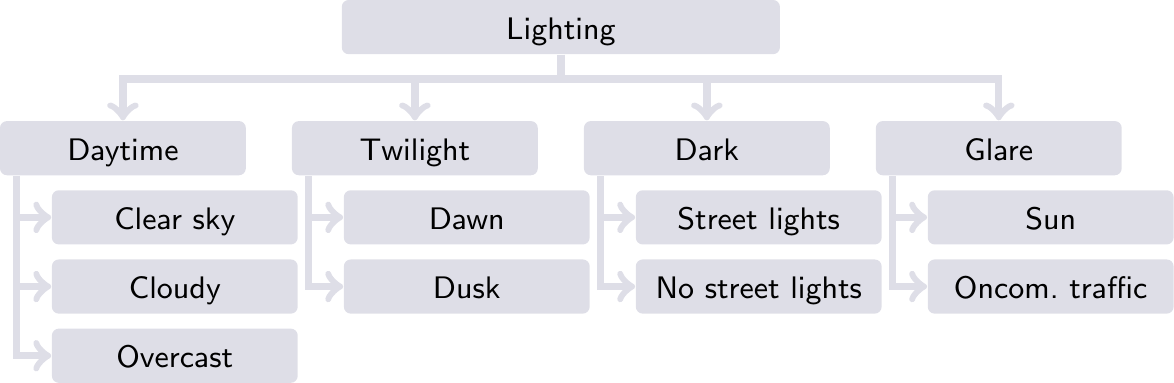}
	\caption{Tags for weather condition, based on \cite{mahmassani2012use} and lighting conditions, see \cite{golob2003relationships}.}
	\label{fig:conditions}
\end{figure*}


%
%

\subsection{Tags for the static environment}
\label{sec:tags selection static}

In this paper, we consider for the static environment the road type (\cref{sec:road type}), the road layout (\cref{sec:road layout}), static objects (\cref{sec:static object}), and a traffic light (\cref{sec:traffic light}).

\subsubsection{Road type}
\label{sec:road type}

The tags for the road type on which the ego vehicle is driving are based on the classification that OpenStreetMaps uses \autocite{HighwayKeyOSM}. We omit the tree of tags because of limited space. We refer the interested reader to \autocite{HighwayKeyOSM}

%

\subsubsection{Road layout}
\label{sec:road layout}

The layout of the road is specified using the tags in \cref{fig:tree road layout}. Here, four categories are defined. Typically, highway roads will mainly be in the category ``Straight''. The second subcategory, i.e., ``Curved'', refers to roads that are highly curved. Typically, the actual speed to safely and comfortably drive these curved roads is lower than the speed limit on the straight section preceding the curved road. For example, a curved road right after a highway exit can often be classified as ``Curved''. The other two categories refer to junctions, whereas ``Pedestrian crossing'' refers to intersections where only a footway is intersecting with the road the ego vehicle is driving on, e.g., a zebra crossing. A large roundabout may be regarded as multiple junctions that are close to each other. For smaller roundabouts, it might be better to treat the roundabout as a whole instead of treating it as multiple junctions. In that case, the ``Roundabout'' tag applies.

\subsubsection{Static object}
\label{sec:static object}

The presence of static objects are described using the tags presented in \cref{fig:tree static object}. A distinction is made between objects that are on the intended path of the ego vehicle and objects that are not on the intended path but are still of importance as they might be blocking the view of the ego vehicle. When a static object is on the intended path of the ego vehicle, the object might be passable - when it is possible to drive over it, or impassable - when the ego vehicle can only avoid undesired interaction with the object by steering around it.

Strictly speaking, every object that is in the field of view of the ego vehicle is blocking part of the ego vehicle's view. For practical reasons, however, an object is classified as ``View blocking'' if the object is significantly blocking parts where it is likely that a traffic participant is present. For example, a building that partially blocks the view of a road is classified as ``View blocking''. For examples of view-blocking objects, see \cite{CATS2015}. A further distinction is made between a parked vehicle or another type of object.

%

\subsubsection{Traffic light}
\label{sec:traffic light}

For a traffic light, we consider the tags ``Red'', ``Amber'', ``Green'', and ``N.A.''. The last tag is applicable in case the traffic light is not operating.
Note that it might be possible that multiple tags are applicable for a scenario. For example, if the traffic light is initially green and turns amber during the timespan of the scenario, both the tags ``Amber'' and ``Green'' are applicable.


\subsection{Tags for the conditions}
\label{sec:conditions}

Separate tags are specified to describe weather and lighting conditions. Weather and lighting conditions are possibly important in the specification of the operational design domain (ODD) of an AV. It might be indicated by an AV developer that the ODD does not include heavy rain or dark night conditions in the absence of street lights. \Cref{fig:conditions} shows tags describing the weather condition (based on \cite{mahmassani2012use}). Tags need to be as specific and quantifiable as possible. Consequently, definitions according to meteorology are followed. 

Tags for different lighting conditions are based on \cite{golob2003relationships}, see \cref{fig:conditions}. Although it might seem straightforward to use the lux level as a quantitative measure for the lighting condition, in this paper, we choose to use a qualitative description, relating the light level to the time of day and the possible presence of artificial lighting. In a study into the influence of ambient lighting conditions on the detection of pedestrians by Automated Emergency Braking systems \autocite{wouters2013influence}, it appeared that lux levels show large variations on the public road. 
The light conditions were measured at a typical junction equipped with street lights during night time. Variations with a factor of 100 to 1000 easily occur due to changes in position underneath a street light. 
Also the presence of other ambient lighting sources has a large influence. As it is not possible to indicate an average lux level, we use a qualitative description of the light level. 
During daytime, there is a strong relation between the weather condition and the available light in a scenario. These weather conditions have been included in the tag tree for lighting.
Glare, a bright and strong light that shines directly onto the ego-vehicle's camera, is another important lighting condition influencing an AV's performance. Glare can be caused by the sun shine while driving to the West just before sunset (or to the East just after sunrise), or by cars in on-coming traffic using high beam headlights. A branch on glare has been added to the lighting tree.

\section{Examples of scenario categories}
\label{sec:examples}

In this section, three different ways are presented to use the tags, presented in \cref{sec:tags}, for defining scenario categories. To illustrate this, we present three scenario categories\footnote{For more examples, we refer the reader to \autocite{degelder2019scenariocategories}.}. 

The first way to define a scenario category using tags is to list the tags that are applicable. To avoid ambiguity, first the top-level tag is mentioned and, next, the lower-level tags that apply. To illustrate this, see the scenario category ``driving on a straight road'' that is schematically shown in \cref{fig:scheme straight}.

\setlength{\figurewidth}{15em}
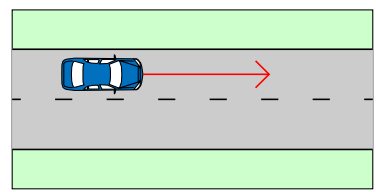
\begin{figure}
	\centering
\begin{tikzpicture}

\definecolor{color0}{rgb}{0.8,1,0.8}
\definecolor{color1}{rgb}{0,0.4375,0.75}

\begin{axis}[
axis background/.style={fill=color0},
height=0.50\figurewidth,
scale only axis,
tick align=outside,
tick pos=left,
ticks=none,
width=\figurewidth,
x grid style={white!69.01960784313725!black},
xmin=-10, xmax=10,
xtick style={color=black},
y grid style={white!69.01960784313725!black},
ymin=-5, ymax=5,
ytick style={color=black}
]
\path [draw=white!80.0!black, fill=white!80.0!black]
(axis cs:-10,2.8)
--(axis cs:10,2.8)
--(axis cs:10,-2.8)
--(axis cs:-10,-2.8)
--cycle;
\addplot [semithick, black]
table {%
-10 2.8
10 2.8
};
\addplot [semithick, black]
table {%
10 -2.8
-10 -2.8
};
\addplot [semithick, black]
table {%
-10 0
-9.52380952380952 0
};
\addplot [semithick, black]
table {%
-7.61904761904762 0
-6.66666666666667 0
};
\addplot [semithick, black]
table {%
-4.76190476190476 0
-3.80952380952381 0
};
\addplot [semithick, black]
table {%
-1.90476190476191 0
-0.952380952380953 0
};
\addplot [semithick, black]
table {%
0.952380952380953 0
1.90476190476191 0
};
\addplot [semithick, black]
table {%
3.80952380952381 0
4.76190476190476 0
};
\addplot [semithick, black]
table {%
6.66666666666667 0
7.61904761904762 0
};
\addplot [semithick, black]
table {%
9.52380952380953 0
10 0
};
\path [draw=black, fill=color1]
(axis cs:-7.25,1.40401785714286)
--(axis cs:-7.24189189189189,1.92633928571429)
--(axis cs:-7.18513513513513,2.12723214285714)
--(axis cs:-7.10405405405405,2.19955357142857)
--(axis cs:-6.97432432432432,2.24776785714286)
--(axis cs:-6.51216216216216,2.30401785714286)
--(axis cs:-3.09864864864865,2.26383928571429)
--(axis cs:-3.00945945945946,2.20758928571429)
--(axis cs:-2.87162162162162,2.03080357142857)
--(axis cs:-2.81486486486487,1.83794642857143)
--(axis cs:-2.75,1.40401785714286)
--(axis cs:-2.75,1.39598214285714)
--(axis cs:-2.81486486486487,0.962053571428571)
--(axis cs:-2.87162162162162,0.769196428571429)
--(axis cs:-3.00945945945946,0.592410714285714)
--(axis cs:-3.09864864864865,0.536160714285714)
--(axis cs:-6.51216216216216,0.495982142857143)
--(axis cs:-6.97432432432432,0.552232142857143)
--(axis cs:-7.10405405405405,0.600446428571428)
--(axis cs:-7.18513513513513,0.672767857142857)
--(axis cs:-7.24189189189189,0.873660714285714)
--(axis cs:-7.25,1.39598214285714)
--cycle;
\path [draw=black, fill=color1]
(axis cs:-4.33108108108108,2.23169642857143)
--(axis cs:-4.33918918918919,2.43258928571429)
--(axis cs:-4.33108108108108,2.48080357142857)
--(axis cs:-4.29864864864865,2.45669642857143)
--(axis cs:-4.25,2.24776785714286)
--cycle;
\path [draw=black, fill=color1]
(axis cs:-4.33108108108108,0.568303571428571)
--(axis cs:-4.33918918918919,0.367410714285714)
--(axis cs:-4.33108108108108,0.319196428571428)
--(axis cs:-4.29864864864865,0.343303571428571)
--(axis cs:-4.25,0.552232142857143)
--cycle;
\path [draw=black, fill=white]
(axis cs:-6.71486486486486,1.40401785714286)
--(axis cs:-6.69864864864865,1.77366071428571)
--(axis cs:-6.65,1.99866071428571)
--(axis cs:-6.58513513513514,2.11919642857143)
--(axis cs:-6.52837837837838,2.12723214285714)
--(axis cs:-6.01756756756757,1.990625)
--(axis cs:-6.06621621621622,1.85401785714286)
--(axis cs:-6.07432432432432,1.64508928571429)
--(axis cs:-6.07432432432432,1.15491071428571)
--(axis cs:-6.06621621621622,0.945982142857143)
--(axis cs:-6.01756756756757,0.809375)
--(axis cs:-6.52837837837838,0.672767857142857)
--(axis cs:-6.58513513513514,0.680803571428571)
--(axis cs:-6.65,0.801339285714285)
--(axis cs:-6.69864864864865,1.02633928571429)
--(axis cs:-6.71486486486486,1.39598214285714)
--cycle;
\path [draw=black, fill=white]
(axis cs:-3.91756756756757,1.40401785714286)
--(axis cs:-3.94189189189189,1.83794642857143)
--(axis cs:-4.03108108108108,2.11116071428571)
--(axis cs:-4.08783783783784,2.16741071428571)
--(axis cs:-4.61486486486486,1.95848214285714)
--(axis cs:-4.56621621621622,1.765625)
--(axis cs:-4.55,1.596875)
--(axis cs:-4.55,1.203125)
--(axis cs:-4.56621621621622,1.034375)
--(axis cs:-4.61486486486486,0.841517857142857)
--(axis cs:-4.08783783783784,0.632589285714286)
--(axis cs:-4.03108108108108,0.688839285714286)
--(axis cs:-3.94189189189189,0.962053571428571)
--(axis cs:-3.91756756756757,1.39598214285714)
--cycle;
\path [draw=black, fill=white]
(axis cs:-6.02567567567568,2.23169642857143)
--(axis cs:-5.81486486486487,2.23169642857143)
--(axis cs:-5.81486486486487,2.07901785714286)
--(axis cs:-5.92027027027027,2.12723214285714)
--cycle;
\path [draw=black, fill=white]
(axis cs:-6.02567567567568,0.568303571428571)
--(axis cs:-5.81486486486487,0.568303571428571)
--(axis cs:-5.81486486486487,0.720982142857143)
--(axis cs:-5.92027027027027,0.672767857142857)
--cycle;
\path [draw=black, fill=white]
(axis cs:-5.76621621621622,2.06294642857143)
--(axis cs:-5.76621621621622,2.19151785714286)
--(axis cs:-5.73378378378378,2.22366071428571)
--(axis cs:-5.20675675675676,2.22366071428571)
--(axis cs:-5.18243243243243,2.18348214285714)
--(axis cs:-5.23108108108108,2.03883928571429)
--(axis cs:-5.30405405405405,2.00669642857143)
--(axis cs:-5.57162162162162,2.02276785714286)
--cycle;
\path [draw=black, fill=white]
(axis cs:-5.76621621621622,0.737053571428571)
--(axis cs:-5.76621621621622,0.608482142857143)
--(axis cs:-5.73378378378378,0.576339285714286)
--(axis cs:-5.20675675675676,0.576339285714286)
--(axis cs:-5.18243243243243,0.616517857142857)
--(axis cs:-5.23108108108108,0.761160714285714)
--(axis cs:-5.30405405405405,0.793303571428571)
--(axis cs:-5.57162162162162,0.777232142857143)
--cycle;
\path [draw=black, fill=white]
(axis cs:-5.15,1.98258928571429)
--(axis cs:-5.02027027027027,2.23169642857143)
--(axis cs:-4.28243243243243,2.23169642857143)
--(axis cs:-4.29054054054054,2.18348214285714)
--(axis cs:-4.70405405405405,2.00669642857143)
--cycle;
\path [draw=black, fill=white]
(axis cs:-5.15,0.817410714285714)
--(axis cs:-5.02027027027027,0.568303571428571)
--(axis cs:-4.28243243243243,0.568303571428571)
--(axis cs:-4.29054054054054,0.616517857142857)
--(axis cs:-4.70405405405405,0.793303571428571)
--cycle;
\path [draw=black, fill=white]
(axis cs:-3.2527027027027,2.24776785714286)
--(axis cs:-3.09054054054054,2.23973214285714)
--(axis cs:-2.96891891891892,2.11919642857143)
--(axis cs:-2.91216216216216,2.02276785714286)
--(axis cs:-2.89594594594595,1.92633928571429)
--(axis cs:-2.89594594594595,1.75758928571429)
--(axis cs:-2.98513513513514,1.91026785714286)
--cycle;
\path [draw=black, fill=white]
(axis cs:-3.2527027027027,0.552232142857143)
--(axis cs:-3.09054054054054,0.560267857142857)
--(axis cs:-2.96891891891892,0.680803571428571)
--(axis cs:-2.91216216216216,0.777232142857143)
--(axis cs:-2.89594594594595,0.873660714285714)
--(axis cs:-2.89594594594595,1.04241071428571)
--(axis cs:-2.98513513513514,0.889732142857143)
--cycle;
\addplot [black]
table {%
-6.63378378378378 2.14330357142857
-6.86081081081081 2.13526785714286
-7.07162162162162 2.07901785714286
-7.13648648648649 1.99866071428571
-7.13648648648649 0.801339285714285
-7.07162162162162 0.720982142857143
-6.86081081081081 0.664732142857143
-6.63378378378378 0.656696428571428
};
\addplot [black]
table {%
-3.00945945945946 1.82991071428571
-2.98513513513514 1.85401785714286
-2.89594594594595 1.71741071428571
-2.89594594594595 1.08258928571429
-2.98513513513514 0.945982142857143
-3.00945945945946 0.970089285714286
};
\addplot [black]
table {%
-3.26081081081081 2.13526785714286
-4.0472972972973 2.16741071428571
-3.00945945945946 1.82991071428571
-3.00945945945946 0.970089285714286
-4.0472972972973 0.632589285714286
-3.26081081081081 0.664732142857143
};
\addplot [semithick, red]
table {%
-2.75 1.4
4.25 1.4
};
\addplot [semithick, red]
table {%
3.5 2.15
4.25 1.4
3.5 0.650000000000001
};
\end{axis}

\end{tikzpicture}
	\caption{Schematic overview for the scenario category ``driving on a straight road''.}
	\label{fig:scheme straight}
\end{figure}
\begin{figure}
	\centering
	\includegraphics{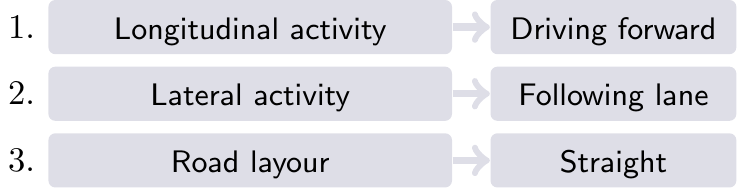}
	\caption{Tags for the scenario category ``driving on a straight road''.}
	\label{fig:tags straight}
\end{figure}

\begin{figure}
	\centering
	\input{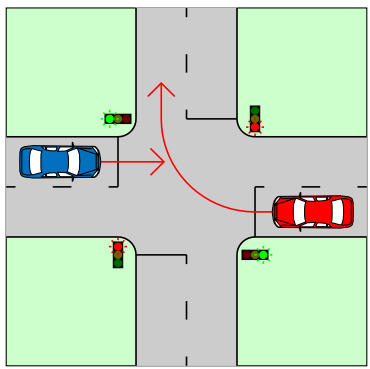}
	\caption{Schematic overview for the scenario category ``oncoming vehicle turns right at signalized junction''. The blue vehicle denotes the ego vehicle.}
	\label{fig:scheme oncoming turning}
\end{figure}

\begin{figure}
	\centering
	\includegraphics{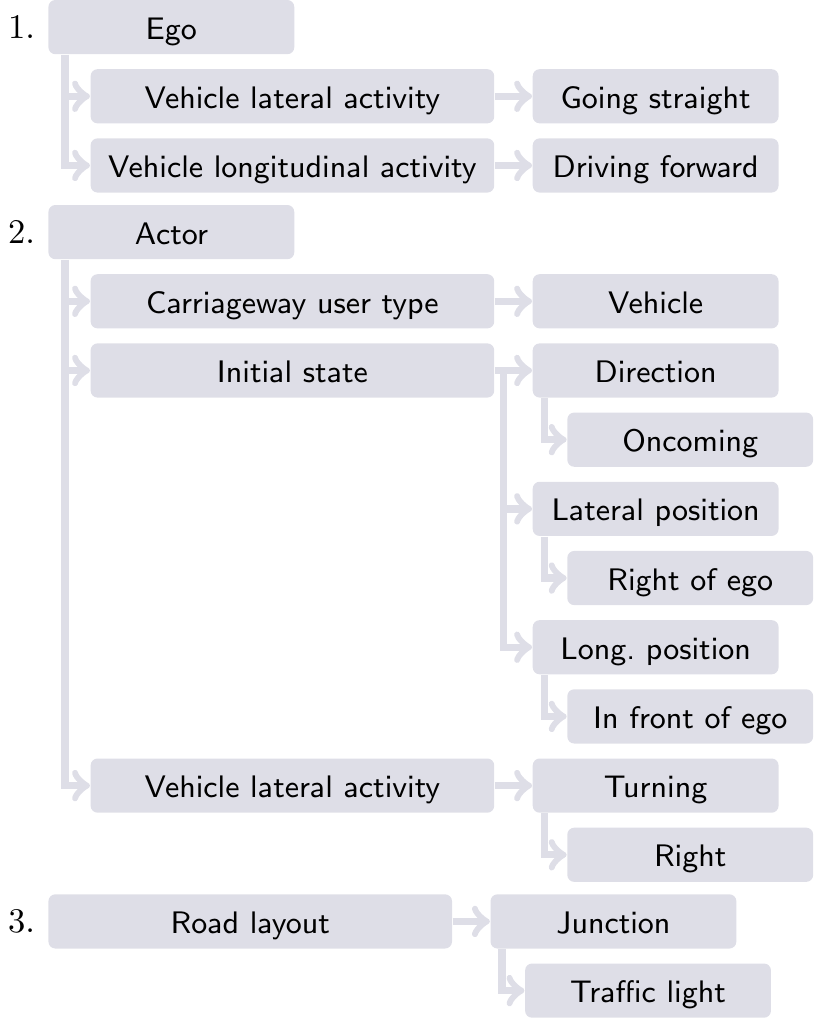}
	\caption{Tags for the scenario category ``oncoming vehicle turns right at signalized junction''.}
	\label{fig:tags oncoming turning}
\end{figure}

It might be that the way to list the tags as shown in \cref{fig:tags straight} is too limited. For example, if multiple vehicles are involved, it might be useful to describe the activities of these vehicles separately. To accommodate this, a selection of tags of a scenario category may be grouped in order to indicate that these tags apply to the same actor. This is illustrated with the example shown in \cref{fig:scheme oncoming turning}: the scenario category ``oncoming vehicle turns right signalized junction''.
The ego vehicle is approaching a junction that is equipped with traffic light signals. The ego vehicle intends to go straight at the crossing. Another vehicle is approaching the junction from the opposite direction. The other vehicle intends to turn right at the junction, such that the trajectories of the other vehicle and the ego vehicle intersect. Note that right-hand traffic is assumed.
In \cref{fig:tags oncoming turning}, the corresponding tags are shown. The first part refers to the ego vehicle that intends to drive straight in forward direction. The second part refers to the other vehicle that turns right. The third part refers describes that the scenario happens at a signalized junction.

\setlength{\figurewidth}{15.0em}

\begin{figure}
	\centering
	\input{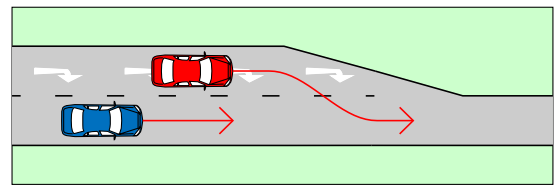}
	\caption{Schematic overview for the scenario category ``cut in at merging lanes''. The blue vehicle denotes the ego vehicle.}
	\label{fig:scheme cut in}		
\end{figure}

\begin{figure}
	\centering
	\includegraphics[width=\linewidth]{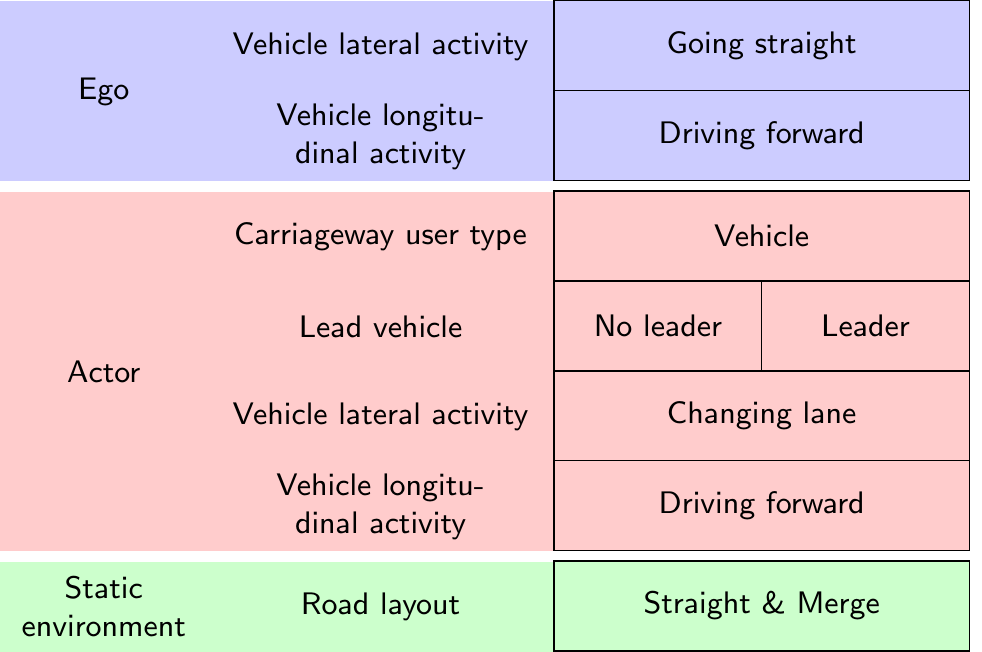}
	\caption{Tags for the scenario category ``cut in at merging lanes''.}
	\label{fig:tags cut in}
\end{figure}

A third way to define a scenario category using tags is presented in case the order in which the tags apply matters. To illustrate this, consider the scenario category ``cut in at merging lanes'' that is schematically shown in \cref{fig:scheme cut in}. 
Another vehicle is driving in the same direction as the ego vehicle in an adjacent lane. The other vehicle makes a lane change because the lanes are merging. In \cref{fig:tags cut in}, the corresponding tags are shown. Most notably, the tag for the ``lead vehicle'' changes from ``no leader'' to ``leader. Note that the direction of the lane change is not described. This could mean that the vehicle could either change lane to the left or to the right. In any case, this actor becomes the lead vehicle, as described my the tag ``lead vehicle''. 

\setlength{\figurewidth}{22.5em}

\section{Conclusions}
\label{sec:conclusions}

The performance assessment of AVs is essential for the legal and public acceptance of AVs as well as for technology development of AVs. 
Scenarios are crucial for the assessment. We proposed a system of tags to group these scenarios into so-called scenario categories. Because the tags are structured in trees, it is possible to define both generic scenario categories and more specific scenario categories.

We have also discussed the relation between scenarios and test cases. For the definition of test cases for the assessment of an AV, it is important to consider the AV's ODD. Currently, there is no unambiguous method known to the authors for the definition of an ODD. A promising method to describe the ODD is using scenario categories. Defining the scenario categories belonging to the AV's ODD according to the proposed method using the system of tags will reduce --- if not remove -- any ambiguity regarding the AV's ODD.

We aimed to provide an extensive list of tags. However, depending on the application, the list of tags might not be extensive enough. Therefore, we will release subsequent versions of the report \autocite{degelder2019scenariocategories} in which more tags might be added.


\bibliographystyle{ieeetran}
\bibliography{bib}

\begin{thebibliography}{10}
\providecommand{\url}[1]{#1}
\csname url@samestyle\endcsname
\providecommand{\newblock}{\relax}
\providecommand{\bibinfo}[2]{#2}
\providecommand{\BIBentrySTDinterwordspacing}{\spaceskip=0pt\relax}
\providecommand{\BIBentryALTinterwordstretchfactor}{4}
\providecommand{\BIBentryALTinterwordspacing}{\spaceskip=\fontdimen2\font plus
\BIBentryALTinterwordstretchfactor\fontdimen3\font minus
  \fontdimen4\font\relax}
\providecommand{\BIBforeignlanguage}[2]{{%
\expandafter\ifx\csname l@#1\endcsname\relax
\typeout{** WARNING: IEEEtran.bst: No hyphenation pattern has been}%
\typeout{** loaded for the language `#1'. Using the pattern for}%
\typeout{** the default language instead.}%
\else
\language=\csname l@#1\endcsname
\fi
#2}}
\providecommand{\BIBdecl}{\relax}
\BIBdecl

\bibitem{bengler2014threedecades}
K.~Bengler, K.~Dietmayer, B.~F{\"a}rber, M.~Maurer, C.~Stiller, and H.~Winner,
  ``Three decades of driver assistance systems: Review and future
  perspectives,'' \emph{IEEE Intelligent Transportation Systems Magazine},
  vol.~6, no.~4, pp. 6--22, 2014.

\bibitem{stellet2015taxonomy}
J.~E. Stellet, M.~R. Zofka, J.~Schumacher, T.~Schamm, F.~Niewels, and J.~M.
  Z\"{o}llner, ``Testing of advanced driver assistance towards automated
  driving: A survey and taxonomy on existing approaches and open questions,''
  in \emph{IEEE 18th International Conference on Intelligent Transportation
  Systems}, 9 2015, pp. 1455--1462.

\bibitem{Helmer2017safety}
T.~Helmer, K.~Kompa{\ss}, L.~Wang, T.~K{\"u}hbeck, and R.~Kates, \emph{Safety
  Performance Assessment of Assisted and Automated Driving in Traffic:
  Simulation as Knowledge Synthesis}.\hskip 1em plus 0.5em minus 0.4em\relax
  Springer International Publishing, 2017, pp. 473--494.

\bibitem{putz2017pegasus}
A.~P\"{u}tz, A.~Zlocki, J.~Bock, and L.~Eckstein, ``System validation of highly
  automated vehicles with a database of relevant traffic scenarios,'' in
  \emph{12th ITS European Congress}, 2017, pp. 1--8.

\bibitem{roesener2017comprehensive}
C.~Roesener, J.~Sauerbier, A.~Zlocki, F.~Fahrenkrog, L.~Wang, A.~V{\'a}rhelyi,
  E.~de~Gelder, J.~Dufils, S.~Breunig, P.~Mejuto, F.~Tango, and J.~Lanati, ``A
  comprehensive evaluation approach for highly automated driving,'' in
  \emph{25th International Technical Conference on the Enhanced Safety of
  Vehicles (ESV)}, 2017.

\bibitem{gietelink2006development}
O.~Gietelink, J.~Ploeg, B.~De~Schutter, and M.~Verhaegen, ``Development of
  advanced driver assistance systems with vehicle hardware-in-the-loop
  simulations,'' \emph{Vehicle System Dynamics}, vol.~44, no.~7, pp. 569--590,
  2006.

\bibitem{wachenfeld2016release}
W.~Wachenfeld and H.~Winner, ``The release of autonomous vehicles,'' in
  \emph{Autonomous Driving}.\hskip 1em plus 0.5em minus 0.4em\relax Springer,
  2016, pp. 425--449.

\bibitem{response2006code}
A.~Knapp, M.~Neumann, M.~Brockmann, R.~Walz, and T.~Winkle, ``Code of practice
  for the design and evaluation of {ADAS},'' \emph{RESPONSE III: a PReVENT
  Project}, 8 2009.

\bibitem{ISO26262}
\BIBentryALTinterwordspacing
{ISO 26262}, \emph{{ISO} 26262: Road {V}ehicles -- {F}unctional {S}afety},
  International Organization for Standardization Std., 2018. [Online].
  Available: \url{https://www.iso.org/standard/68383.html}
\BIBentrySTDinterwordspacing

\bibitem{elrofai2018scenario}
\BIBentryALTinterwordspacing
H.~Elrofai, J.-P. Paardekooper, E.~de~Gelder, S.~Kalisvaart, and O.~Op~den
  Camp, ``Scenario-based safety validation of connected and automated
  driving,'' Netherlands Organization for Applied Scientific Research, TNO,
  Tech. Rep., 2018. [Online]. Available:
  \url{http://publications.tno.nl/publication/34626550/AyT8Zc/TNO-2018-streetwise.pdf}
\BIBentrySTDinterwordspacing

\bibitem{degelder2020ontology}
E.~de~Gelder, J.-P. Paardekooper, A.~Khabbaz~Saberi, H.~Elrofai, O.~Op~den
  Camp, J.~Ploeg, L.~Friedman, and B.~De~Schutter, ``Ontology for scenarios for
  the assessment of automated vehicles,''
  \emph{https://arxiv.org/abs/2001.11507}, 2020.

\bibitem{sae2018j3016}
{SAE International}, ``Taxonomy and definitions for terms related to driving
  automation systems for on-road motor vehicles,'' Tech. Rep. J3016, 6 2018.

\bibitem{ISO34502}
\BIBentryALTinterwordspacing
{ISO 34502}, \emph{{ISO} 34502: Road {V}ehicles -- {E}ngineering Framework and
  Process of Scenario-Based Safety Evaluation}, International Organization for
  Standardization Std., 2020. [Online]. Available:
  \url{https://www.iso.org/standard/78951.html}
\BIBentrySTDinterwordspacing

\bibitem{degelder2019completeness}
E.~de~Gelder, J.-P. Paardekooper, O.~Op~den Camp, and B.~De~Schutter, ``Safety
  assessment of automated vehicles: How to determine whether we have collected
  enough field data?'' \emph{Traffic Injury Prevention}, vol.~20, no.~S1, pp.
  162--170, 2019.

\bibitem{smith2007tagging}
G.~Smith, \emph{Tagging: People-Powered Metadata for the Social Web}.\hskip 1em
  plus 0.5em minus 0.4em\relax New Riders Publishing, 2007.

\bibitem{craft2004tagging}
D.~H. Craft, P.~A. Caro, J.~Pasqua, and D.~C. Brotsky, ``Tagging data assets,''
  U.S. Patent US 6,704,739 B2, 3 9, 2004.

\bibitem{kong2017joint}
Q.~Kong, Y.~Xu, W.~Wang, and M.~D. Plumbley, ``A joint detection-classification
  model for audio tagging of weakly labelled data,'' in \emph{2017 IEEE
  International Conference on Acoustics, Speech and Signal Processing
  (ICASSP)}, 2017, pp. 641--645.

\bibitem{ellis2011semantic}
K.~Ellis, E.~Coviello, and G.~R. Lanckriet, ``Semantic annotation and retrieval
  of music using a bag of systems representation,'' in \emph{International
  Society for Music Information Retrieval}, 2011, pp. 723--728.

\bibitem{voss2006collaborative}
J.~Voss, ``Collaborative thesaurus tagging the {W}ikipedia way,'' \emph{arXiv
  preprint cs/0604036}, 2006.

\bibitem{degelder2020scenariomining}
E.~de~Gelder, J.~Manders, C.~Grappiolo, J.-P. Paardekooper, O.~Op~den Camp, and
  B.~De~Schutter, ``Real-world scenario mining for the assessment of automated
  vehicles,'' in \emph{IEEE International Transportation Systems Conference
  (ITSC)}, 2020, pp. 1073--1080.

\bibitem{opdencamp2014cats}
O.~Op~den Camp, A.~Ranjbar, J.~Uittenbogaard, E.~Rosen, R.~Fredriksson, and
  S.~de~Hair, ``Overview of main accident scenarios in car-to-cyclist accidents
  for use in {AEB}-system test protocol,'' in \emph{International Cycling
  Safety Conference}, 2014.

\bibitem{USDoT2007precrashscenarios}
W.~G. Najm, J.~D. Smith, and M.~Yanagisawa, ``Pre-crash scenario typology for
  crash avoidance research,'' U.S. Department of Transportation Research and
  Innovative Technology Administration, Tech. Rep. DOT HS 810 767, 4 2007.

\bibitem{lenard2014typical}
J.~Lenard, A.~Badea-Romero, and R.~Danton, ``Typical pedestrian accident
  scenarios for the development of autonomous emergency braking test
  protocols,'' \emph{Accident Analysis \& Prevention}, vol.~73, pp. 73--80,
  2014.

\bibitem{lara2019harmonized}
A.~Lara, J.~Skvarce, H.~Feifel, M.~Wagner, and T.~Tengeiji, ``Harmonized
  pre-crash scenarios for reaching global vision zero,'' in \emph{26th
  International Technical Conference on the Enhanced Safety of Vehicles (ESV)},
  no. 19-0110, 2019.

\bibitem{molloy2017dynamic}
S.~Molloy, T.~Ramos, and S.~Thakar, ``Dynamic hierarchical tagging system and
  method,'' U.S. Patent US 9,613,099 B2, 4 4, 2017.

\bibitem{Bonnin2014}
S.~Bonnin, T.~H. Weisswange, F.~Kummert, and J.~Schmuedderich, ``General
  behavior prediction by a combination of scenario-specific models,''
  \emph{IEEE Transactions on Intelligent Transportation Systems}, vol.~15,
  no.~4, pp. 1478--1488, 8 2014.

\bibitem{ploeg2018cetran}
J.~Ploeg, E.~de~Gelder, M.~Slav{\'i}k, E.~Querner, T.~Webster, and N.~de~Boer,
  ``Scenario-based safety assessment framework for automated vehicles,'' in
  \emph{16th ITS Asia-Pacific Forum}, 2018, pp. 713--726.

\bibitem{elrofai2016scenario}
H.~Elrofai, D.~Worm, and O.~Op~den Camp, ``Scenario identification for
  validation of automated driving functions,'' in \emph{Advanced Microsystems
  for Automotive Applications 2016}.\hskip 1em plus 0.5em minus 0.4em\relax
  Springer, 2016, pp. 153--163.

\bibitem{roesener2016scenariobased}
C.~Roesener, F.~Fahrenkrog, A.~Uhlig, and L.~Eckstein, ``A scenario-based
  assessment approach for automated driving by using time series classification
  of human-driving behaviour,'' in \emph{IEEE 19th International Conference on
  Intelligent Transportation Systems (ITSC)}, 11 2016, pp. 1360--1365.

\bibitem{deGelder2017assessment}
E.~de~Gelder and J.-P. Paardekooper, ``Assessment of automated driving systems
  using real-life scenarios,'' in \emph{IEEE Intelligent Vehicles Symposium
  (IV)}, 2017, pp. 589--594.

\bibitem{UNECE2011consolidated}
{World Forum for Harmonization of Vehicle Regulations}, ``Consolidated
  resolution on the construction of vehicles (r.e.3), revision 2,'' UNECE,
  Economic and Social Council, Tech. Rep. ECE/TRANS/WP.29/78/Rev.2, 2011.

\bibitem{degelder2019scenariocategories}
\BIBentryALTinterwordspacing
E.~de~Gelder, O.~Op~den Camp, and N.~de~Boer, ``Scenario categories for the
  assessment of automated vehicles,'' CETRAN, Tech. Rep., 2020, version 1.7.
  [Online]. Available:
  \url{http://cetran.sg/wp-content/uploads/2020/01/REP200121_Scenario_Categories_v1.7.pdf}
\BIBentrySTDinterwordspacing

\bibitem{mahmassani2012use}
H.~Mahmassani, R.~Mudge, T.~Hou, and J.~Kim, ``Use of mobile data for
  weather-responsive traffic management models,'' U.S. Department of
  Transportation, Tech. Rep., 2012.

\bibitem{golob2003relationships}
T.~F. Golob and W.~W. Recker, ``Relationships among urban freeway accidents,
  traffic flow, weather, and lighting conditions,'' \emph{Journal of
  Transportation Engineering}, vol. 129, no.~4, pp. 342--353, 2003.

\bibitem{HighwayKeyOSM}
\BIBentryALTinterwordspacing
OpenStreetMaps. Key:highway. Accessed: April 2018. [Online]. Available:
  \url{https://wiki.openstreetmap.org/wiki/Key:highway}
\BIBentrySTDinterwordspacing

\bibitem{CATS2015}
O.~Op~den Camp, S.~de~Hair, E.~de~Gelder, and I.~Cara, ``Observation dtudy into
  the influence of a view-blocking obstruction at an intersection on bicycle
  and passenger car velocity profiles,'' in \emph{International Cyclist Safety
  Conference}, 2015.

\bibitem{wouters2013influence}
R.~Wouters and O.~Op~den Camp, ``Influence of light conditions on detection of
  pedestrians by {AEB} systems, report for the {N}etherlands ministry of
  infrastructure and the environment,'' TNO Integrated Vehicle Safety, Tech.
  Rep. TNO 2013 R11962, 2013.

\end{thebibliography}


%


\end{document}